# On the mathematics of beauty: beautiful images

A. M. Khalili[a]*

[a] *School of Creative Arts and Engineering, Staffordshire University, United Kingdom*

ABSTRACT

The question of beauty has inspired philosophers and scientists for centuries. Today, the study of aesthetics is an active research topic in many fields. In this paper, we will study the simplest kind of beauty which can be found in simple visual patterns. The proposed approach shows that aesthetically appealing patterns deliver a higher amount of information over multiple levels in comparison with less aesthetically appealing patterns when the same amount of energy is used. The proposed approach is used to classify aesthetically appealing patterns.

## 1. Introduction

The study of aesthetics started with the work of ancient Greek, and today it is an active research topic in many fields. Predicting the aesthetic appeal of images is beneficial for many applications, such as recommendation and retrieval in multimedia systems. The development of a model of aesthetic judgment is also a major challenge in evolutionary art [8], [9], where only images with high aesthetic quality should be generated. The development of the social media and the fast growth in visual media content have increased the requirement of aesthetic assessment systems.

In this paper, a novel approach to classify aesthetically appealing images will be presented. The main contribution of this paper is showing that aesthetically appealing patterns deliver a higher amount of information over multiple levels in comparison with less aesthetically appealing patterns when the same amount of energy is used.

## 2. Related work

Datta et al. [10] extracted 56 visual features from an image and used them to train a statistical model to classify the images as“beautiful” or “ugly”. Some examples of the used features include: mean pixel intensity, relative color frequencies, mean pixel hue, and mean pixel saturation. They also used photographic rules of thumb such as the rule-of-thirds. Other features related to aspect ratio, texture, and low depth-of-field were also used. Ke et al. [11] used features that describe the spatial distribution of color, edges, brightness, and blur. Aydin et al. [12] computed perceptually calibrated ratings for a set of meaningful and fundamental aesthetic attributes such as depth, sharpness, tone, and clarity, which together form an “aesthetic signature” of the image. Other works have also investigated the role of photographic composition [13], [14], [15], [16], colour compatibility [17], [18], [19], and the use of other features such as object types in the scene [20].

Recently, convolutional neural networks (CNNs), which can automatically learn the aesthetic features, have been applied to the aesthetic quality assessment problem [21], [22], [23], [24], promising results were reported.

This work is more related to the information theory based approaches. Birkhoff [25] proposed an information theory approach to aesthetic, he used a mathematics based aesthetic measure, where the measure of aesthetic quality is in a direct relation to the degree of order O, and in a reverse relation to the complexity C, $M = O/C$. Eysenck [26], [27], [28] argued that the aesthetic measure has to be in a direct relation to the complexity rather than an inverse relation $M = O \times C$. Javid et al. [29] conducted a survey on the use of entropy to quantify order and complexity, they also proposed a computational measure of complexity, their measure is based on the information gain from specifying the spatial distribution of pixels and their uniformity and non-uniformity. Herbert Franke [30] argued that artists should provide an information flow of about 16 bits/sec for their works to be perceived as aesthetically appealing and harmonious. Datasets such as [31], [32], [33], [34] and [35] are collected from communities where images are uploaded and scored in response to photographic challenges. The main limitation of these datasets is that the images are very rich, diverse, and highly subjective, which will make the aesthetic assessment process very complicated. Therefore, the datasets in [43] and [36] will be used to test the proposed approach. The datasets contain simple visual patterns generated by the same physical process.

## 3. Proposed approach

In this section, the simplest kind of beauty that can be found in simple visual patterns will be studied. The images of the dataset in [43] shown in Fig. 11 and the images of the dataset in [36] will be used, the dataset in [36] contains two groups of images, the first one is “more aesthetically appealing” images Fig. 1, and the second one is "less aesthetically appealing" images Fig. 2. These two groups are rated by ten persons. The ten persons were asked to determine if each pattern is beautiful or not. If the score (the number of persons who selected the pattern as beautiful) is higher than the average score, then the pattern belongs to the first group, otherwise it belongs to the second group.

* Corresponding author. e-mail: a.m.khalili@outlook.com

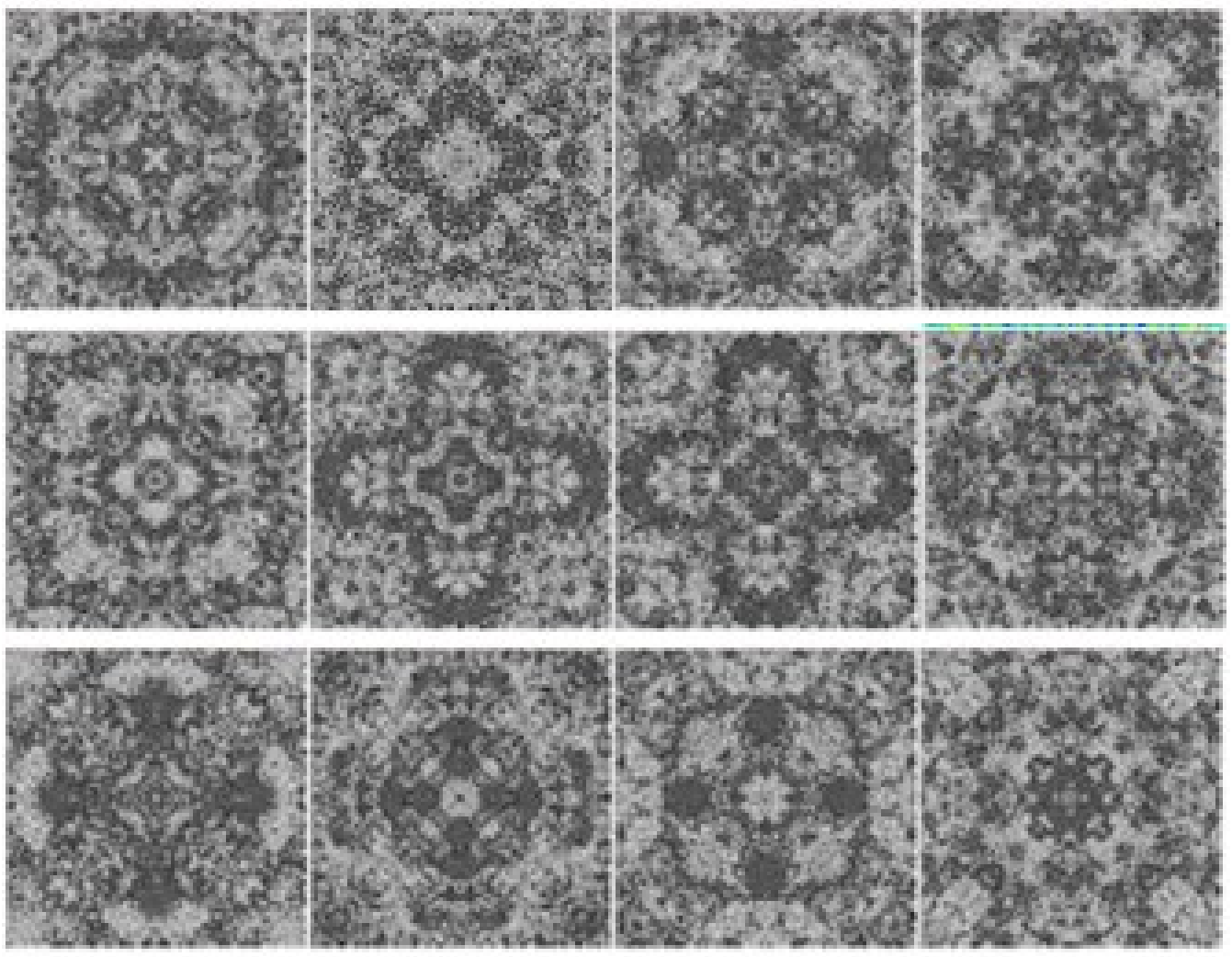

Fig. 1. Images in the first group.

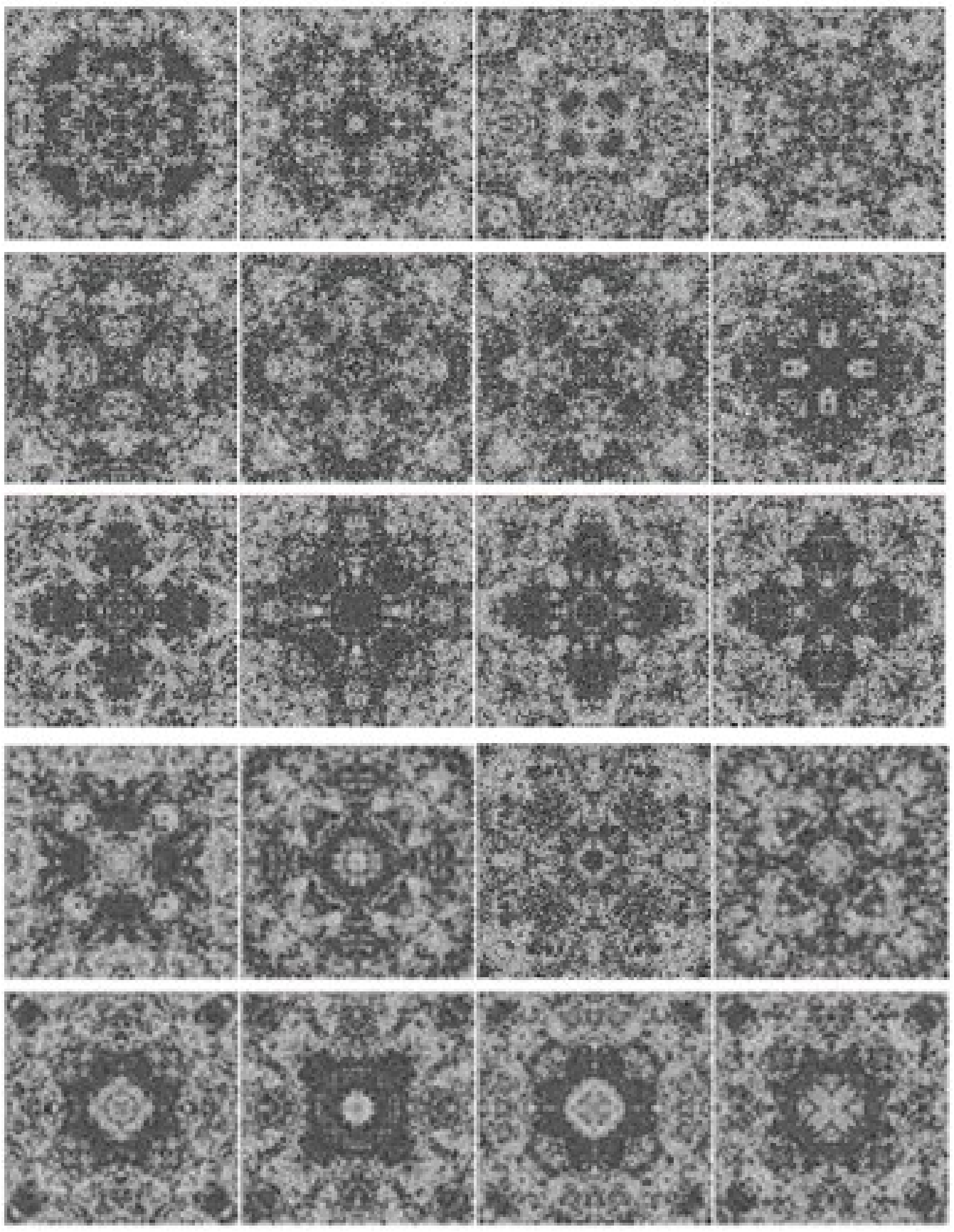

Fig. 2. Images in the second group.

To analyze the images of Fig.1 and Fig. 2, if we start from the center of the image to the boundary, we notice that the number of transitions between lighter and darker values is larger for images in Fig. 1, furthermore; the intensity of the transitions is higher. This will result in increasing the high-energy part of the distribution of the gradient of the image. Moreover, we notice that the high-energy part of the distributions of the images of Fig.1 is larger than the high energy part of the distributions of the images in Fig. 2 when both have the same amount of energy, and since the most part of the distribution is located in the low energy region, this means that increasing the high energy part of the distribution will increase the entropy.

The basic idea of the proposed approach is that aesthetically appealing patterns have a balance between randomness and regularity, and aesthetically appealing patterns are those which are closer to this optimal point. The entropy and energy will be used as measures of this balance. The resulted distribution of this optimization process can be uniquely identified by maximizing the entropy giving that the energy levels are constant and the total energy is constant.

Fig.4 shows the distribution of the gradient of one image in the dataset, the same distribution has shown up for all the images in the dataset. We can observe the similarity between the resulted distribution and the Maxwell-Boltzmann distribution which is shown in Fig. 3. Furthermore, using the above analysis, our problem now is exactly the same problem that Boltzmann [37] solved to derive the distribution of the energies of gas particles at equilibrium. Boltzmann argued that the Maxwell-Boltzmann distribution [38, 39] is the most probable distribution and it will arise by maximizing the multiplicity (which is the number of ways the particles can be arranged) giving that the number of particles is constant as described by (1), the energy levels that the particles can take are constant as described by (2), and the total energy is constant as described by (3). The multiplicity is given by (4), and the entropy is given by (5).

$$\sum_{i} n_i = Constant \quad (1)$$

$$\varepsilon_1, \varepsilon_2, \dots, \varepsilon_n \;\; Constant \quad (2)$$

$$\text{Energy} \;= \sum_{i} n_i \varepsilon_i = Constant \quad (3)$$

$$\Omega = \frac{N!}{n_1! n_2! \dots n_n!} \quad (4)$$

$$\text{Entropy} \;= \log(\Omega) \quad (5)$$

Where N is the total number of particles, $n_i$ is the number of particles at the $\varepsilon_i$ energy level. Maximizing the entropy is equivalent to maximizing the multiplicity. By taking ln(Ω) we get

$$\ln(\Omega) \;= \ln(N!) - \sum_{i} \ln(n_i!) \qquad (6)$$

Using Stirling approximation, we get

$$\ln(\Omega) = \mathrm{N} \ln(N) - N - \sum_{i} [n_i \ln(n_i) - n_i] \qquad (7)$$

The Maxwell-Boltzmann distribution gives the number of particles at each energy level. Using the Lagrange multiplier method to maximize the entropy using the constraints in (1), (2), and (3) we get

$$n_i = e^{-\alpha-\beta\varepsilon_i} \quad (8)$$

Where $\alpha, \beta$ are the Lagrange multipliers. The distribution in 3D and 2D spaces can be written in the form given by (9) and (10) respectively, and the distribution is shown in Fig.3.

$$f(v) = \left(\frac{m}{2\pi kT}\right)^{\frac{3}{2}} 4\pi v^2\, e^{-\frac{mv^2}{2kT}} \quad (9)$$

$$f(v) = \left(\frac{m}{2\pi kT}\right) 2\pi v\, e^{-\frac{mv^2}{2kT}} \quad (10)$$

Where v is the speed of the particle, m is the mass of the particle, T is the temperature and k is Boltzmann constant.

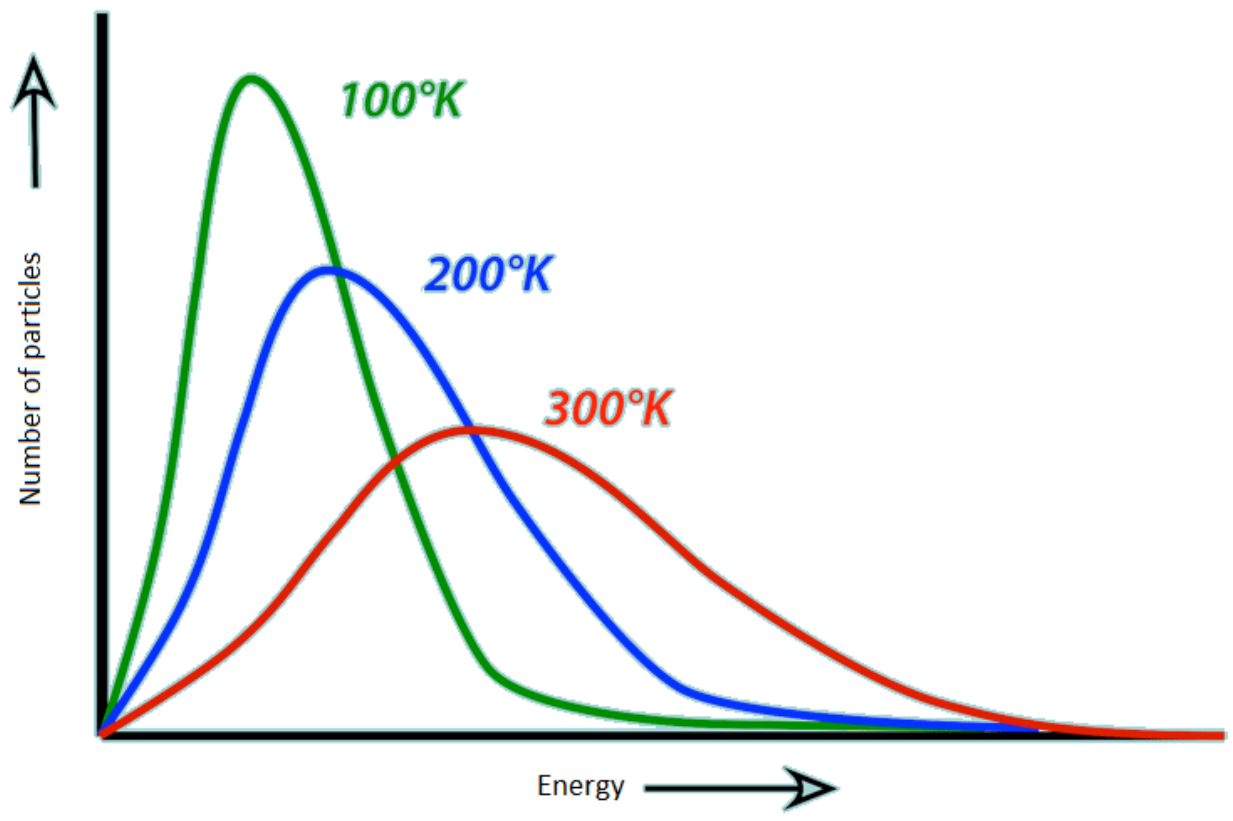

Fig. 3. The Maxwell-Boltzmann distribution for different temperature values.

Similarly, in our problem, the energy levels $\varepsilon_1, \varepsilon_2, \ldots, \varepsilon_n$ are the values which the pixels can take, they will be 0, 1, 2, …, 255 for grayscale images. These energy levels must be constant as described in (11), $n_i$ is the number of pixels at the energy level $\varepsilon_i$, the total number of pixels should also be constant as described in (12). Finally, the total energy which is given by (13) must also be constant.

$$\varepsilon_1, \varepsilon_2, \ldots, \varepsilon_n \; Constant \; (11)$$

$$\sum_{\mathrm{i}} n_i = Constant \; (12)$$

$$\text{Energy } = \sum_{\mathrm{i}} n_i \varepsilon_i = Constant \; (13)$$

The constraints given in (11), (12), and (13) are exactly the same constraints used by Boltzmann to derive the Maxwell-Boltzmann distribution, and by maximizing the entropy, the same distribution given by (8)-(10) will arise. Maximizing the entropy will result in a flat distribution; however, the constant energy constraint will produce a balance between order and randomness. Maximizing the entropy using constant energy can then be seen as delivering the highest possible amount of information using the same amount of energy. Fig. 4 shows the distribution of the gradient of an image in the dataset. Fig. 5 shows the distribution of the gradient of the gradient of the same image.

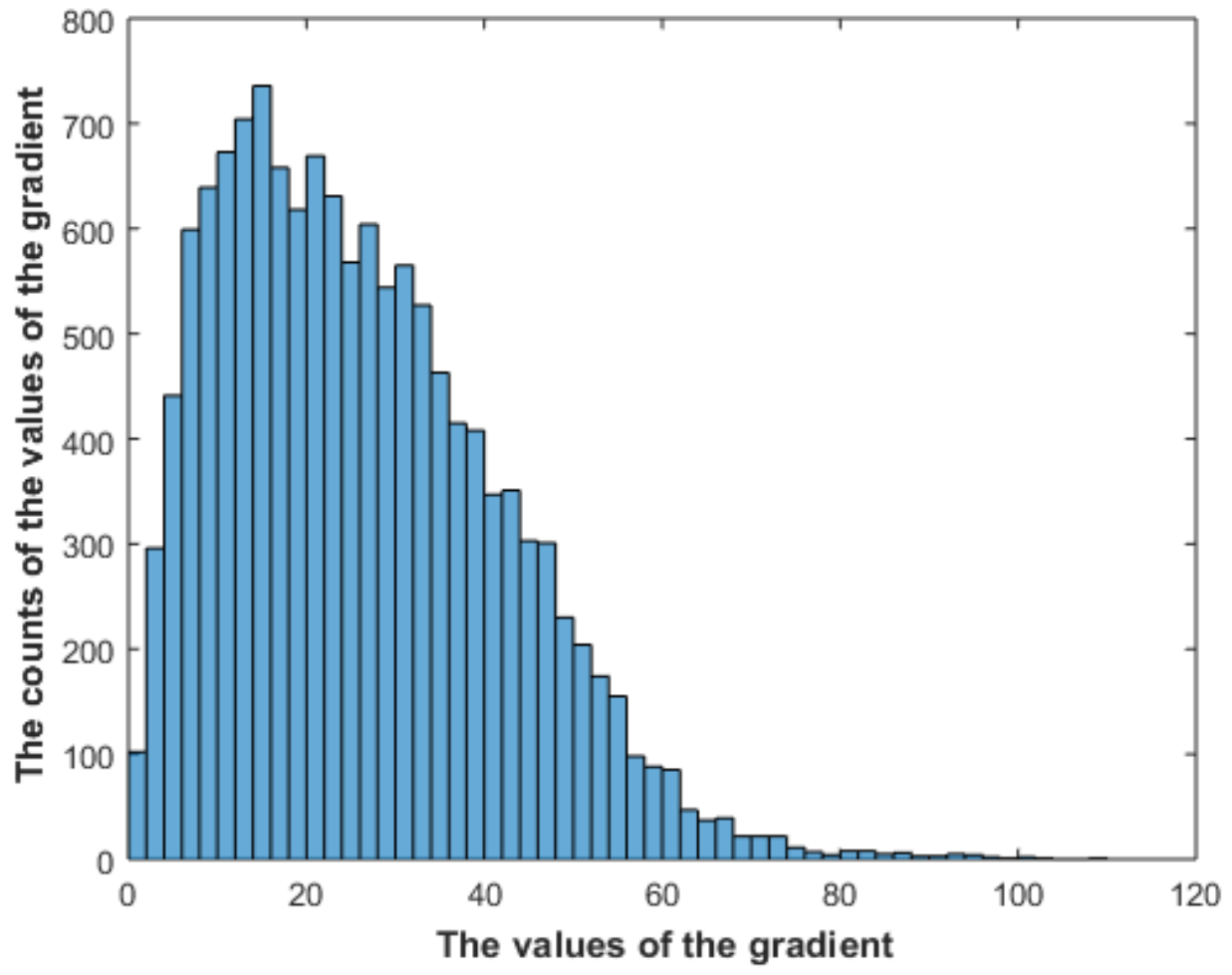

Fig. 4. The distribution of the gradient of one image in the dataset.

The same distribution has appeared for all the gradient of the images, and the gradient of the gradient of the images, which may suggest that the same law must be satisfied at each level. The multiple levels approach will be used to cope with energy and entropy limitation in representing the spatial arrangement of the pattern. Due to the complexity of the structure of the visual patterns, the gradient over multiple levels will be used to represent the spatial arrangement of the visual patterns, where the first level represents the image, the second level represents the gradient of the image, and the third level represents the gradient of the gradient of the image. The measures of aesthetic quality M states that the sum of the entropies of the three levels should be maximum. The measure is given by (14)

$$\mathrm{M} = \sum_{\mathrm{i}} \mathrm{Entropy(Li)} \qquad (14)$$

$L_1$ is the image, $L_2$ is the gradient of the image, and $L_3$ is the gradient of the gradient of the image. Entropy is Shannon entropy, and the energy of the three levels must be the same. Fig. 6 shows the M values of images in Fig. 1 and Fig. 2 along with other images in the same category.

However, comparing images that have the same energy at each level is rather limited; furthermore, the above analysis does not say anything about the relation between the energies of different levels. Fig. 7 shows the sum of the distances between the energies of different levels for images in Fig. 1 and Fig. 2 along with other images in the same category.

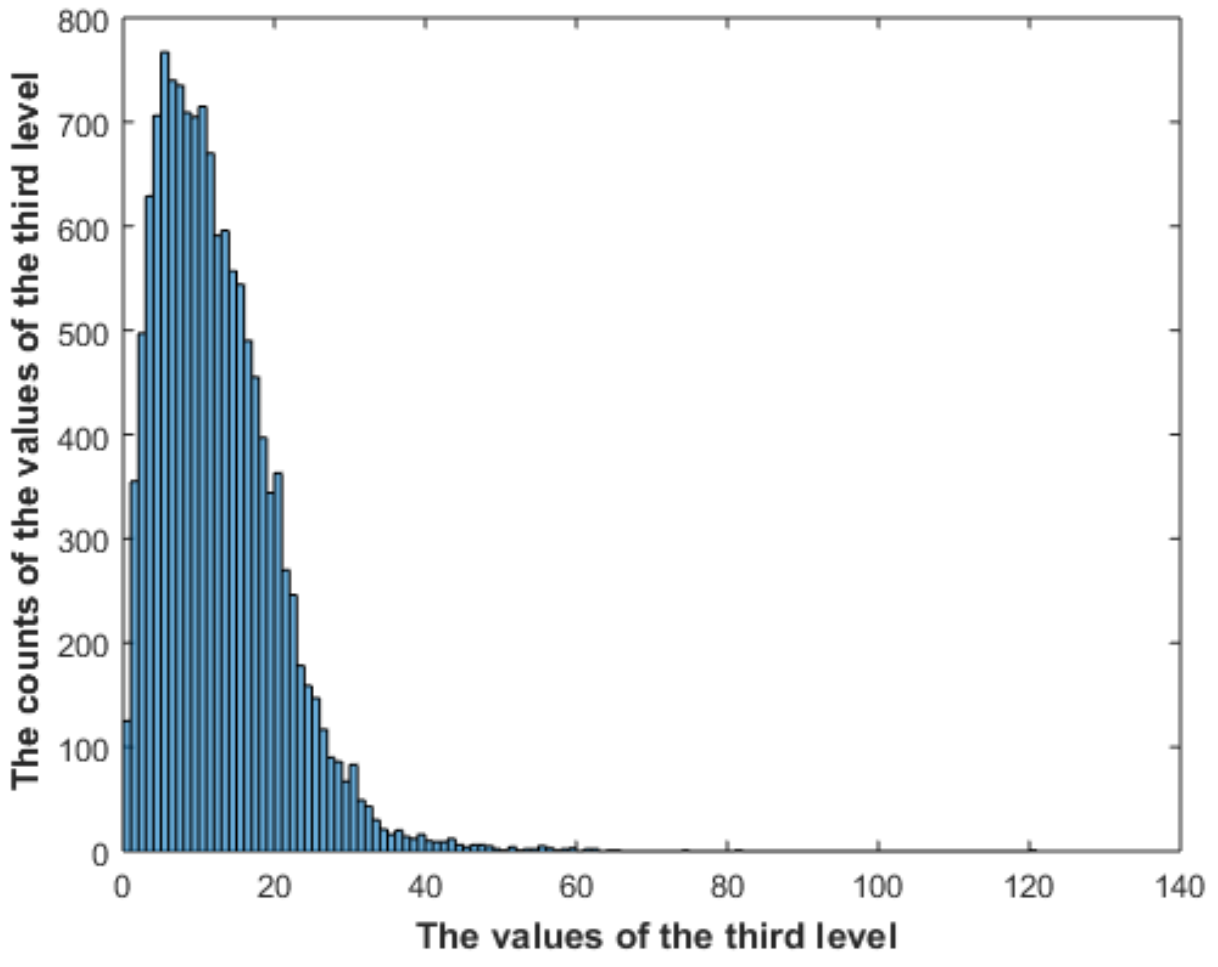

Fig. 5. The distribution of the gradient of the gradient of one image in the dataset.

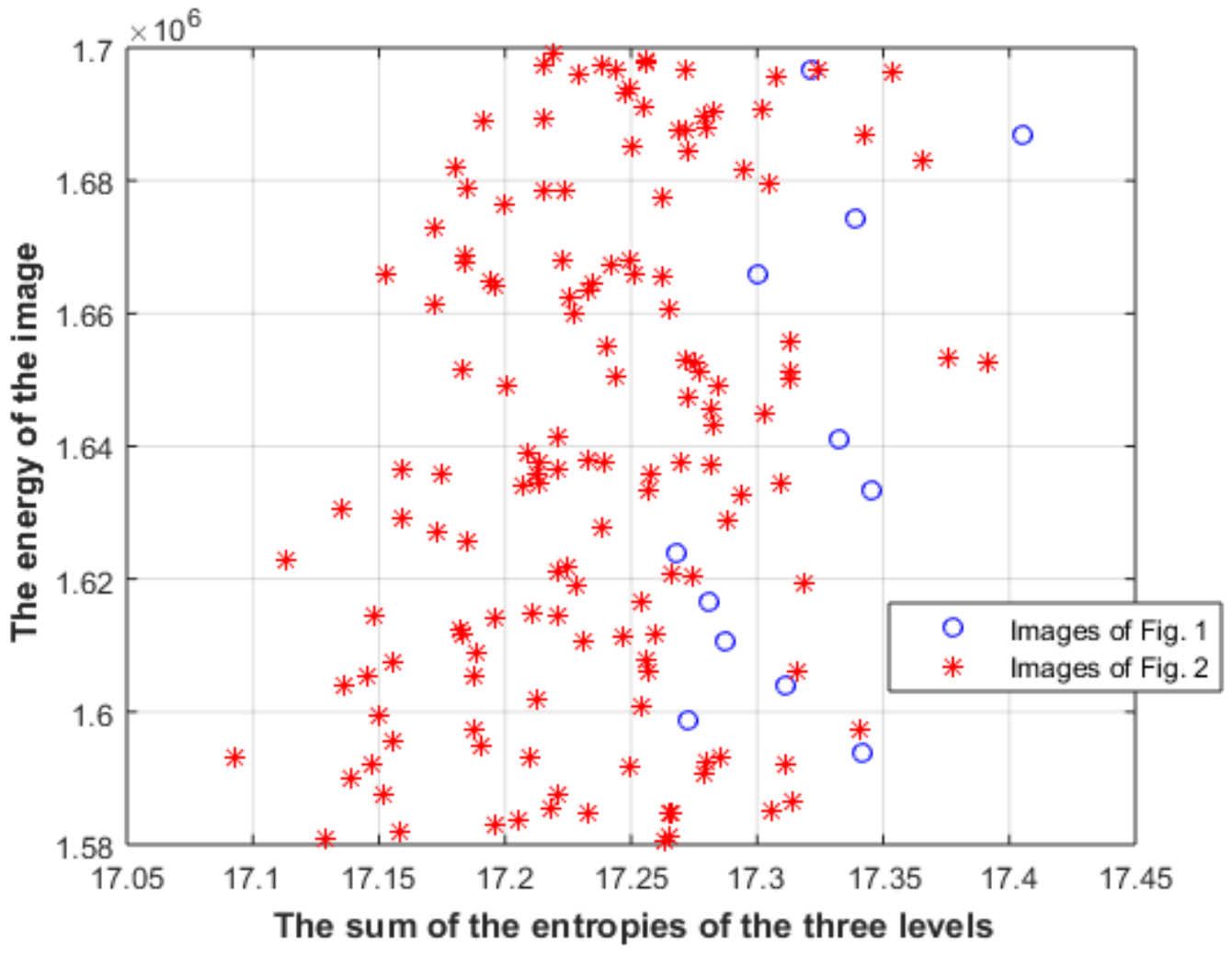

Fig. 6. The M values of images in Fig.1 and Fig.2.

The blue circles represent the images of Fig. 1, and the red stars represent the images of Fig. 2 along with other images in the same category. The distances of aesthetically appealing images are different from the distances of the less aesthetically appealing images. To relax the above constraint and to be able to compare images that have the same first level energy only, the aesthetically appealing images at different energy levels of Fig. 1 are used as reference images, and the distances between the energies of the tested image should be as close as possible to the distances of the reference image $R_i$ as described by (15), furthermore; the equation described by (14) should be also satisfied. In other words, M should be maximized and Md should be minimized.

$$\mathrm{Md} = |\textstyle\sum_i \mathrm{Distance(Ri)} - \sum_i \mathrm{Distance(Li)}| \qquad (15)$$

Where Distance($R_i$) is the distance between the energy of the ith level and the energy of the i+1 level, and the energy of the first level only should be the same. The metrics will be calculated on the center part of the image since it gets most of the attention, where 20 pixels from each side of the image will be neglected. Fig. 8 shows the combination of the two metrics where the sum of the entropies and the energies of the three levels is shown after scaling each energy and entropy to value between 0 and 1.

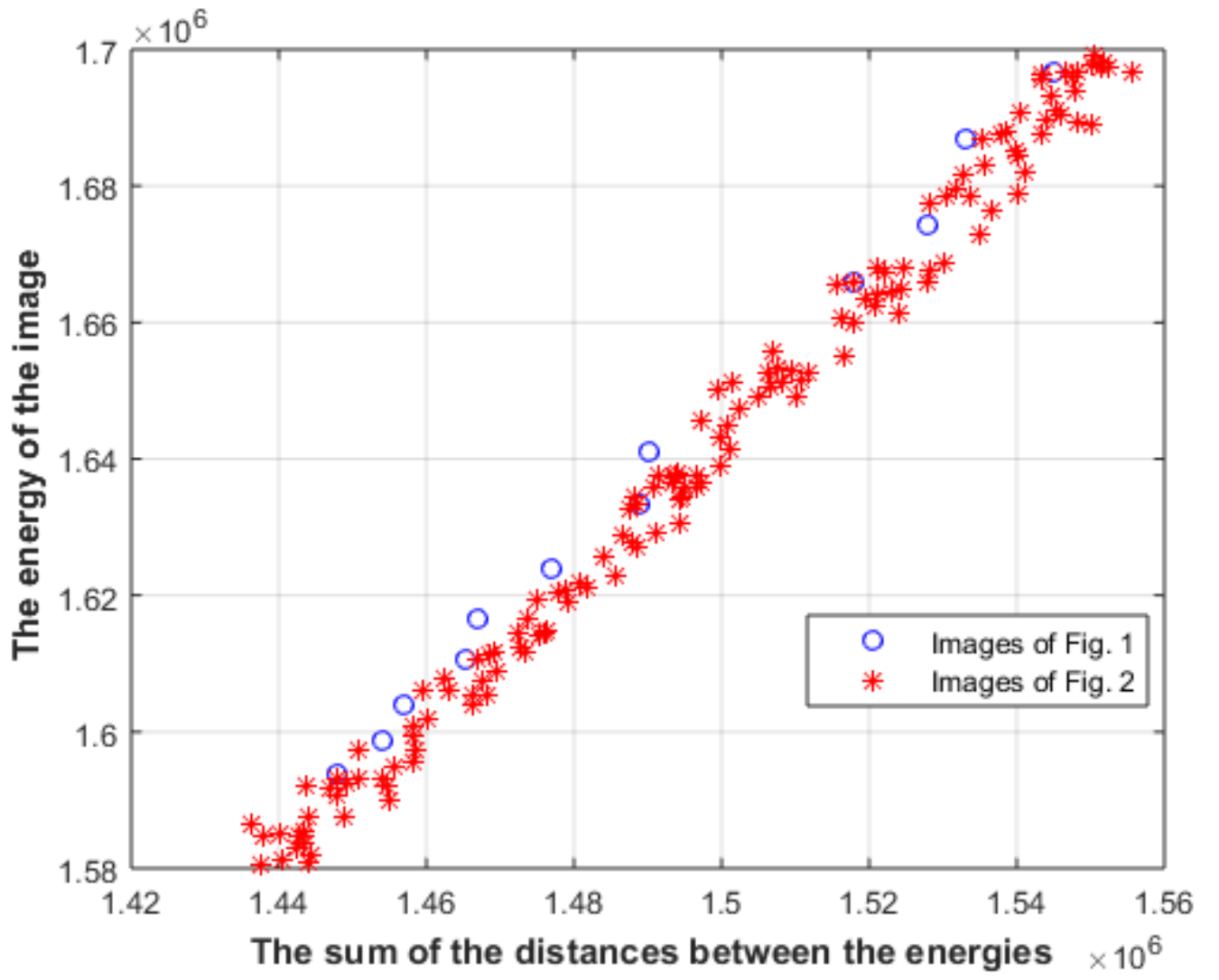

Fig. 7. The sum of the distances between the energies of different levels for images in Fig.1 and Fig.2.

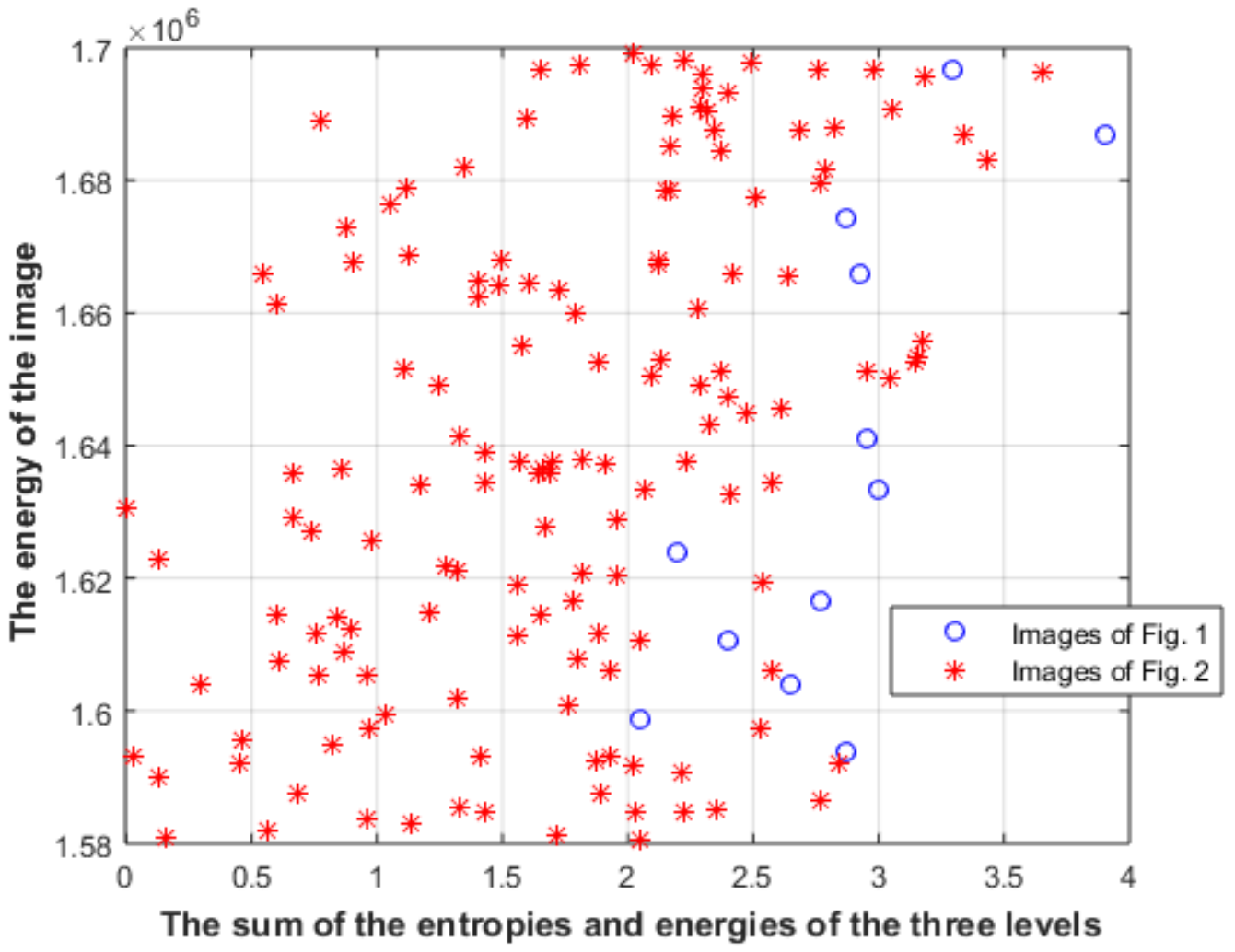

Fig. 8. The sum of the entropies and energies of the three levels of images in Fig.1 and Fig.2.

## 4. Results

Due to the small number of images in the two datasets. The proposed approach can not be compared to deep learning based approaches. The proposed approach will be compared with three related approaches, the first one is based on Birkhoff model [40], [41], where Shannon entropy and image compressibility are used to represent the order and complexity of Birkhoff model. Fig. 9 shows the Shannon entropy and image compressibility for the images of Fig. 1 and Fig. 2. The results show that the two groups of images cannot be easily classified using this approach.

Then, the proposed approach will be compared with an approach based on Benford law [42], where the histogram of the image is compared with the histogram described by Benford law.

Fig. 10 shows the difference between the histograms of the images of Fig. 1 and Fig. 2, and the histogram described by Benford law. The results also show that the two groups of images cannot be easily classified using this approach.

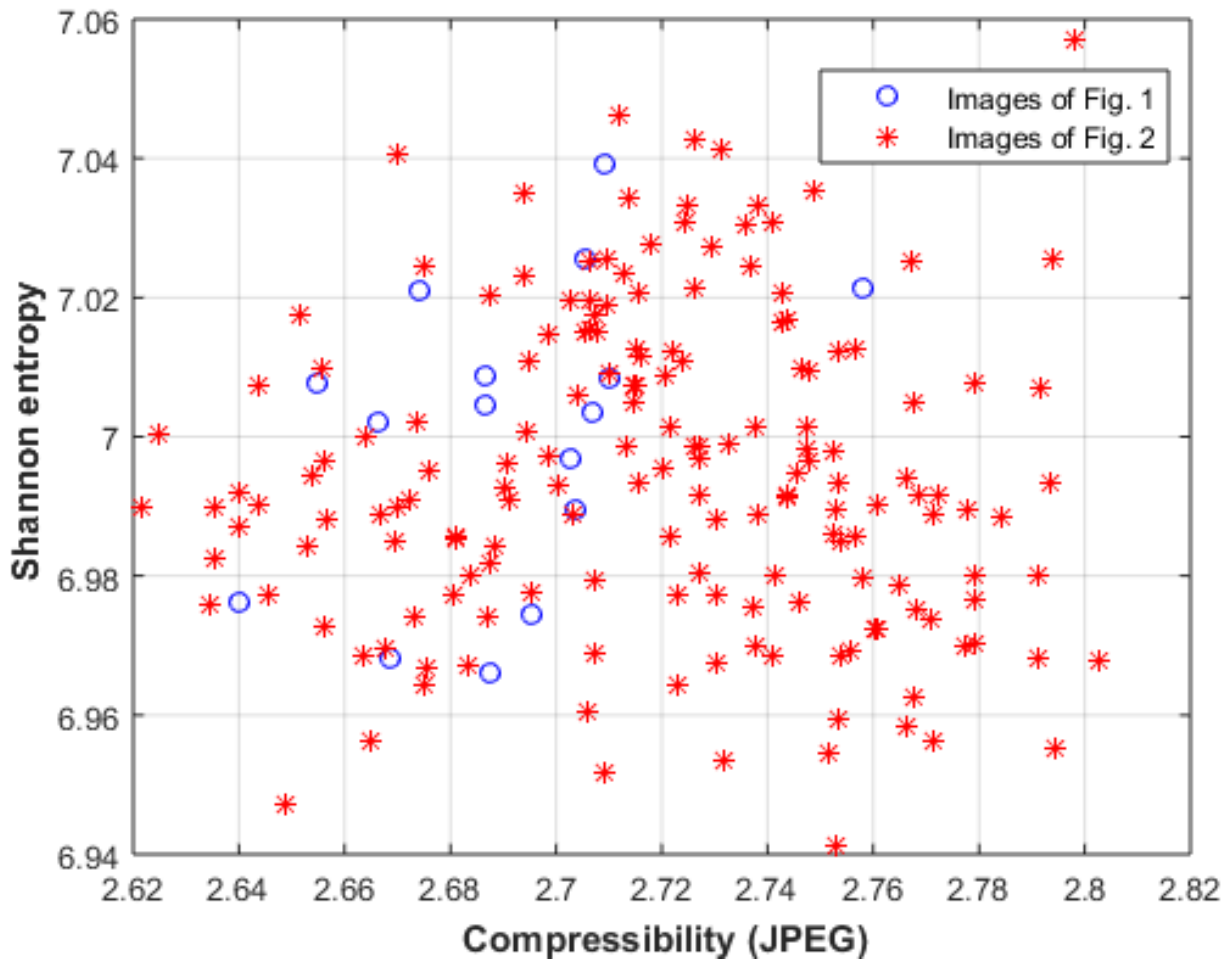

Fig. 9. Shannon entropy vs image compressibility for images of Fig. 1 and Fig. 2.

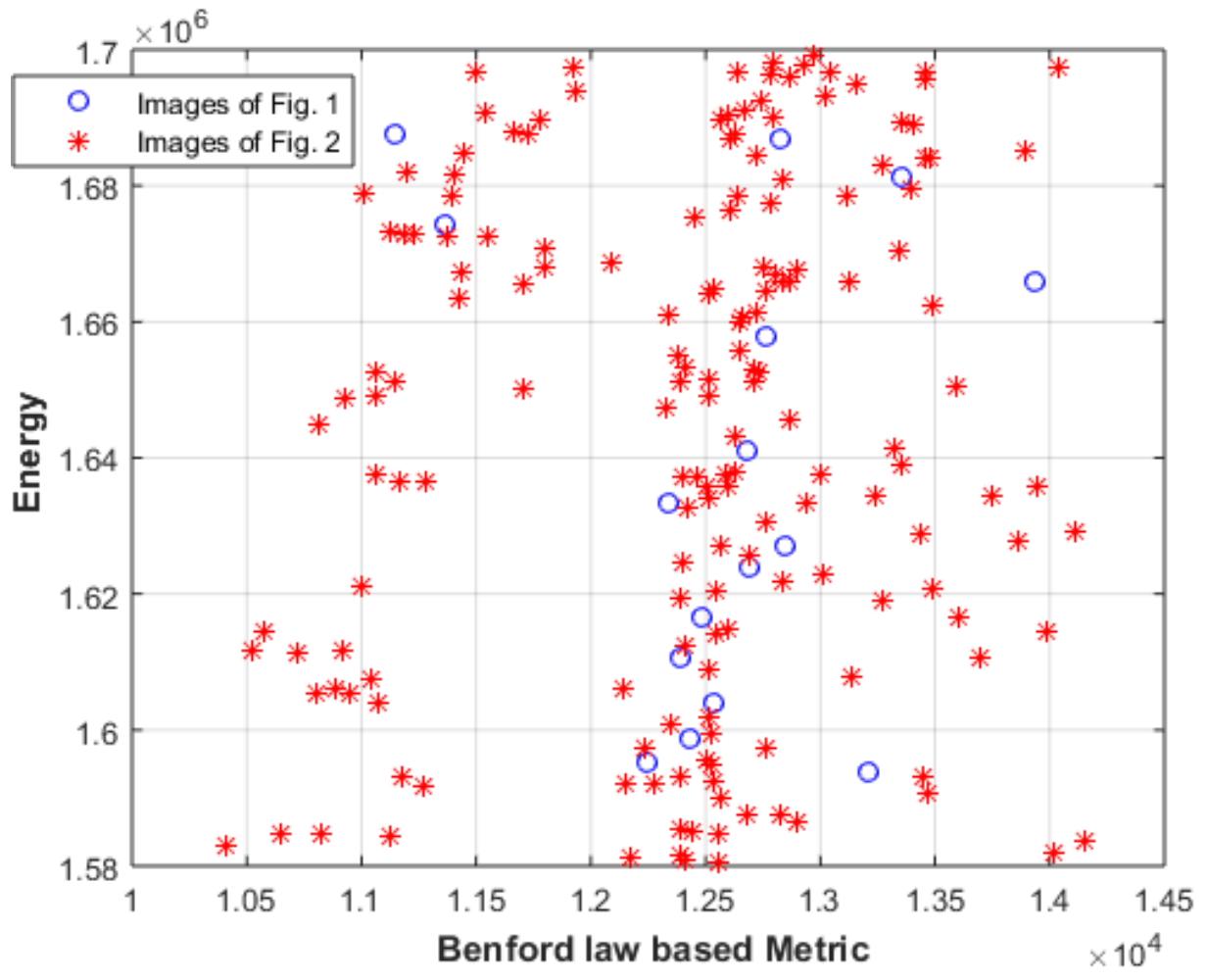

Fig. 10. The difference between the histograms of the images of Fig. 1 and Fig. 2, and the histogram described by Benford law.

To further test the proposed approach, we will test it on the dataset proposed in [43], [44]. Fig. 11 shows the patterns of the set, the first two lines represent asymmetrical patterns; the last two lines represent symmetrical patterns. Fifty-five persons rated the patterns, the patterns start from not beautiful (left) to beautiful (right) line by line. The number next to each pattern in Fig. 12, Fig. 13, and Fig 14 represents the line number and the position of the pattern in the line (starting from left to right). For instance, 43 is the third pattern in line four.

Fig. 12 shows the energy and the entropy of the first level, the results show that the symmetrical patterns of line 3 and line 4 have higher entropy than the asymmetrical patterns when the same energy is used. This matches with the rating given by the fifty-five persons and with several studies [45-48] that showed consistent preferences for symmetry. The patterns 41, 42, and 43 have roughly the same energy, but the entropy of 43 is larger than the entropy of 42, which is larger than the entropy of 41.

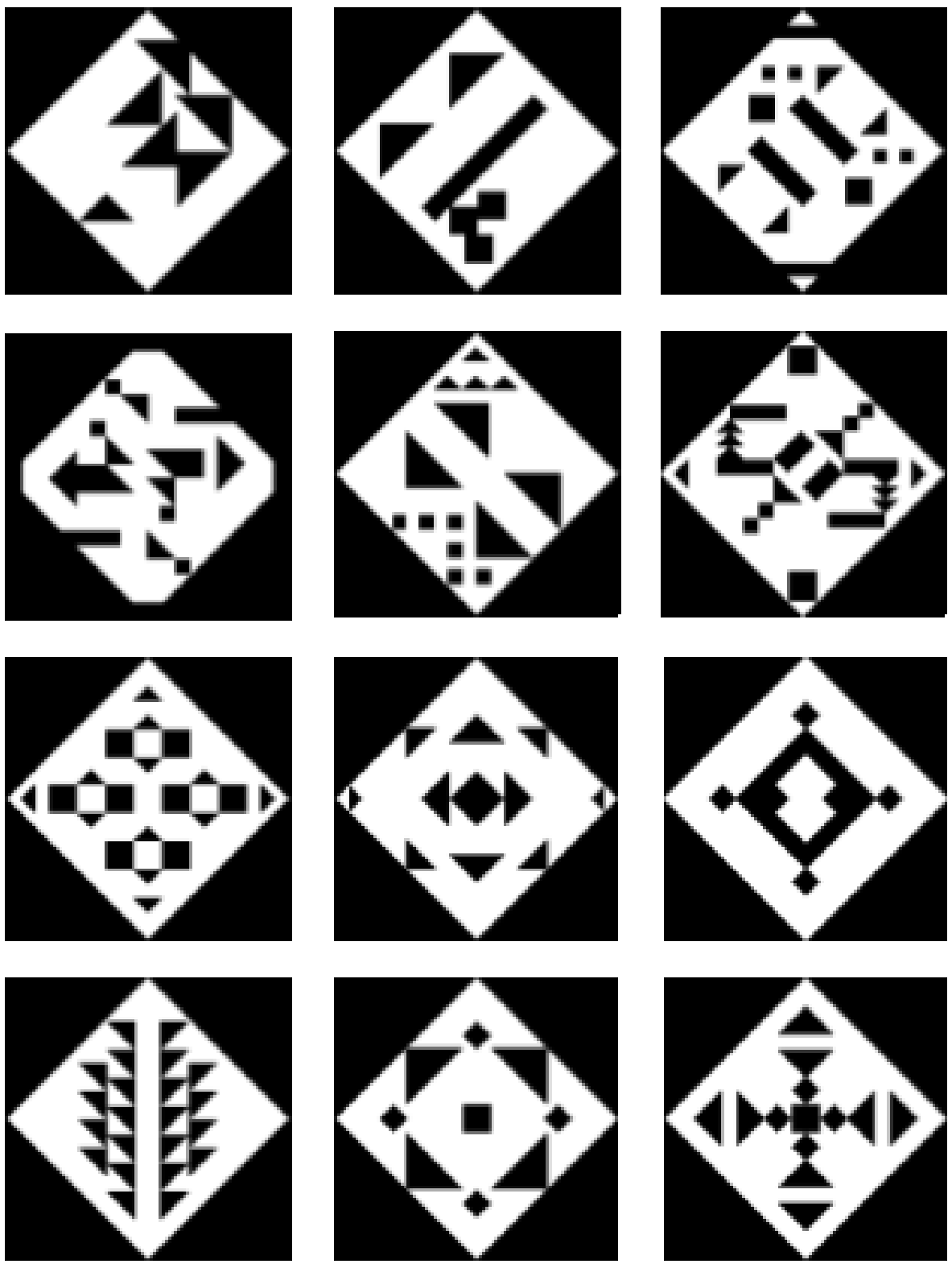

**Fig. 11.** Patterns from the set proposed in [43, 44], ordered from not beautiful (left) to beautiful (right) line by line.

Fig. 13 shows the sum of the entropies of the first two levels, again the symmetrical patterns of line 3 and line 4 have higher sum than the other patterns when the same energy is used. For instance, the patterns 13, 32, and 33 have roughly the same energy, but the sum of 33 is larger than the sum of 32, which is larger than the sum of 13. This also matches with the rating of the Fifty-five persons. We can also see that the patterns 11 and 21 have lower sum than the other patterns.

Fig. 14 shows the distance between the energies of the first two levels. The symmetrical patterns of line 3 and line 4 have lower distance than the other patterns when the same energy is used. For instance, the patterns 13, 32, and 33 have roughly the same energy, but the distance of 33 is lower than the distance of 32, which is lower than the distance of 13. The patterns 41, 42, and 43 also have roughly the same energy, but the distance of 43 is lower than the distance of 42; however, 42 has higher distance than 41. We can also see that the patterns 11 and 21 have higher distance than the other patterns. These results show a close match with the rating given by the Fifty-five persons.

Fig. 15 shows the results of applying an information gain based approach proposed in [29] on the images in Fig. 11. The results show a link between information gain and empirical aesthetic judgement in the case of asymmetrical patterns but not for symmetrical patterns

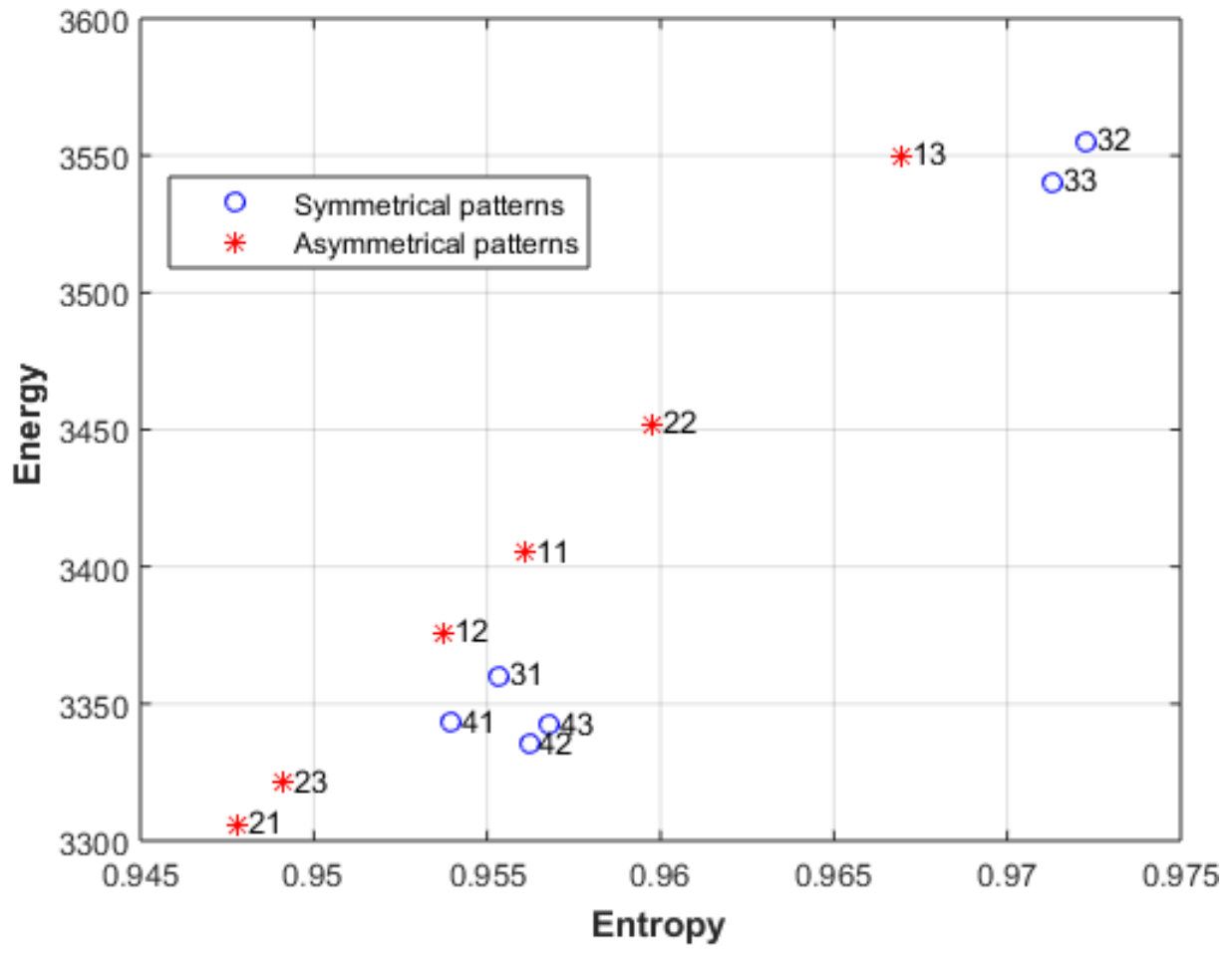

**Fig. 12.** The energy and the entropy of the first level of the images in Fig.11.

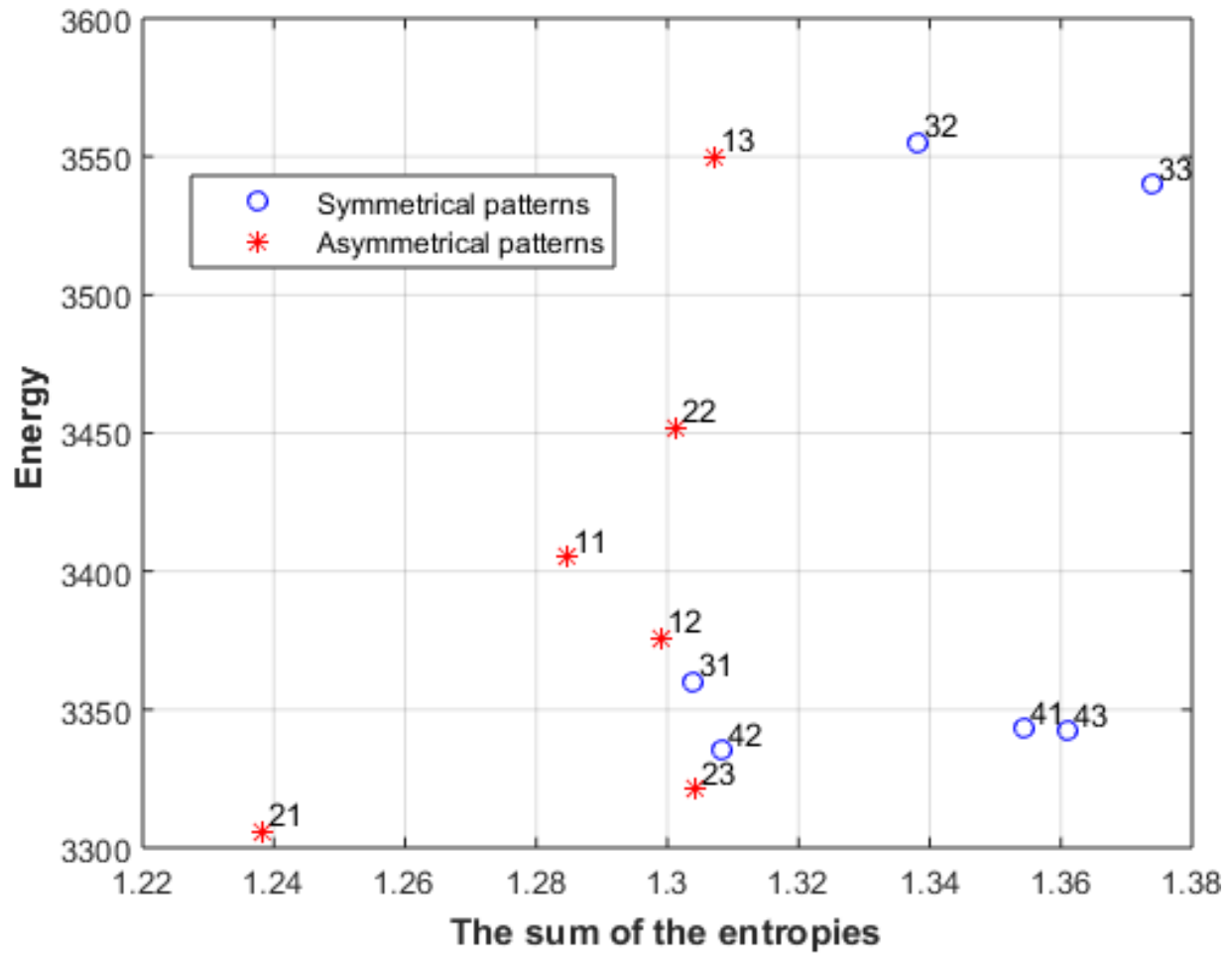

**Fig. 13.** The sum of the entropies of the first two levels of the images in Fig.11.

To give a more intuitive analysis for the results, we will take two extreme cases, the first one is an image with only one color, and the second one is an image with equal probabilities for all colors. The first case will produce a distribution of one pulse at one energy level, while the second case will produce a flat distribution. In the case of music the first case will give a piece with only one note repeated many times, and the second case will produce a piece with all possible notes, in both cases no aesthetically appealing patterns will be produced, where the first pattern will be too regular and the second one will be too random. The aesthetically appealing patterns represent a balance between these two extreme cases, and the closer we get to the Maxwell-Boltzmann distribution, the higher the aesthetic score of the pattern. Now if we take one aesthetically appealing pattern and rearrange the pixels randomly, we will get a random pattern that has the same distribution; however, the gradient of this random pattern will produce a distribution closer to the flat distribution than the gradient of the original pattern. Similarly, if we arrange the aesthetically appealing pattern such that the pixels with the same values are close to each other, the gradient of the resulted pattern will produce a distribution closer to a pulse than the gradient of the original pattern, and again the distribution of the gradient of aesthetically appealing patterns represents a balance between these two extreme cases, and the closer we get to the Maxwell-Boltzmann distribution, the higher the aesthetic score of the pattern.

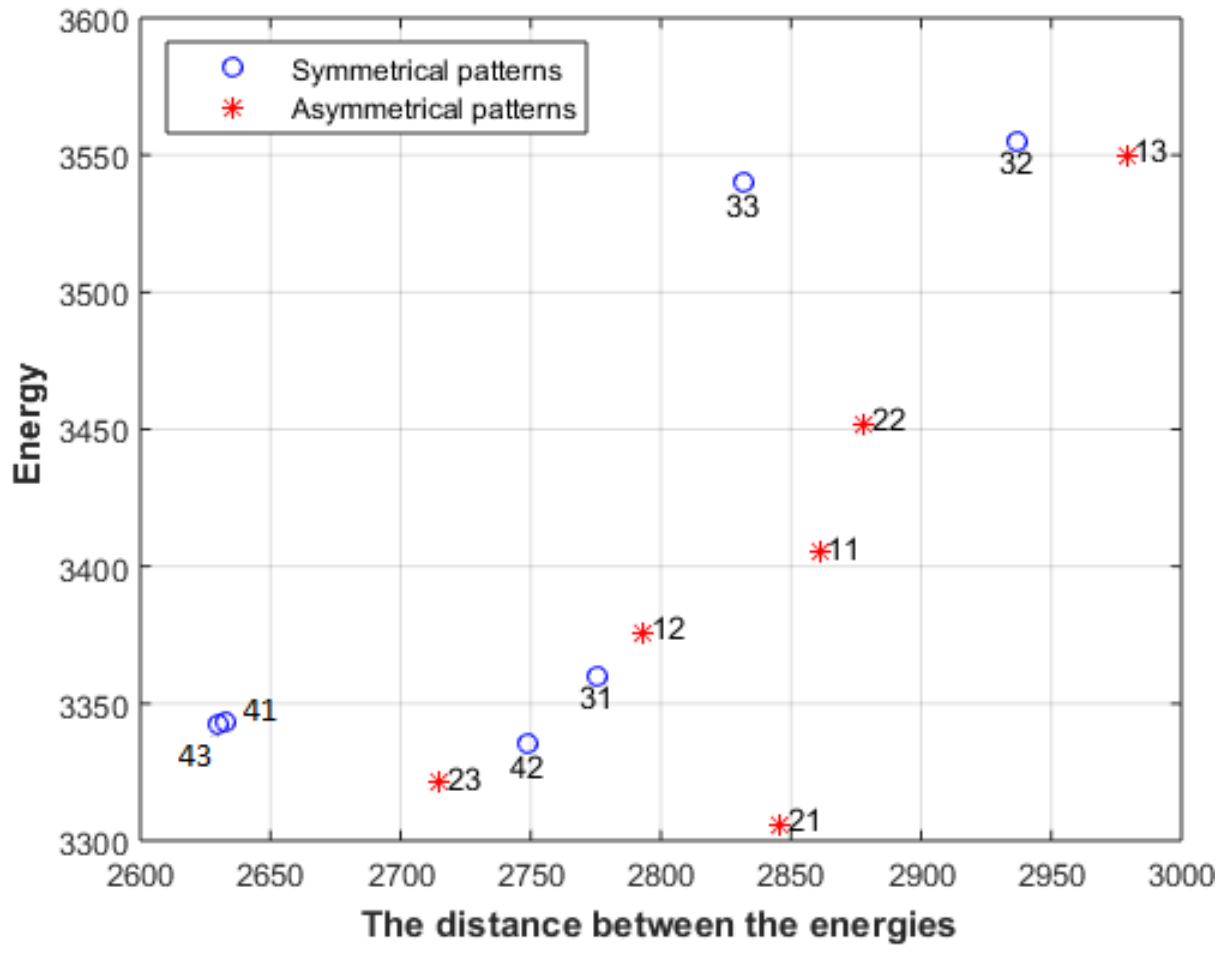

**Fig. 14.** The distance between the energies of the first two levels of the images in Fig. 11.

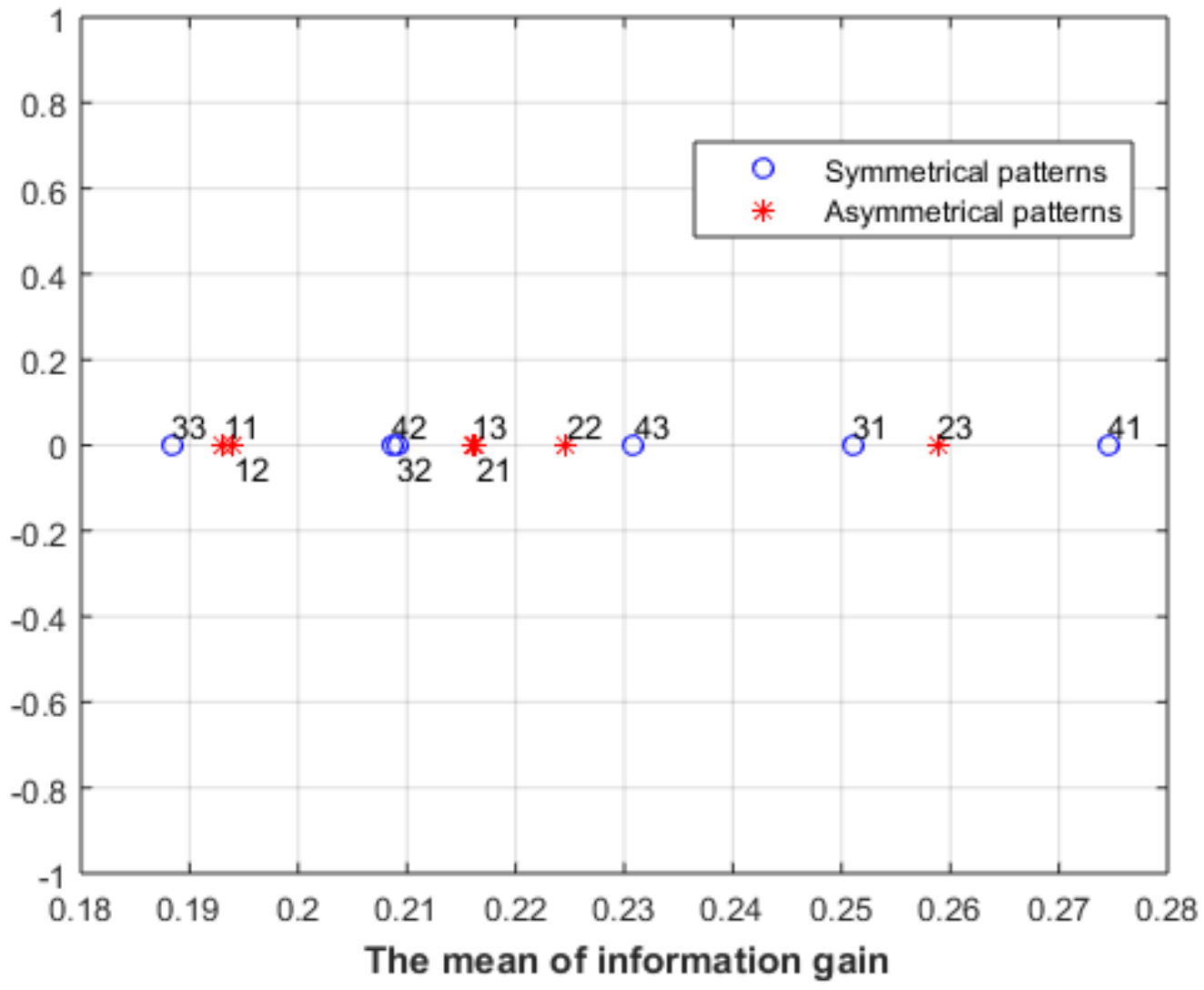

**Fig. 15.** The mean information gain of the images in Fig. 11.

These results show that the proposed model is more accurate in classifying aesthetically appealing visual patterns. The results show that aesthetically appealing patterns deliver a higher amount of information in comparison with less aesthetically appealing patterns when the same amount of energy is used, the results also show that the distances between the energies of the levels are different for the aesthetically appealing patterns. One limitation of the proposed approach is that few aesthetically appealing patterns show lower M value and higher Md value than the less aesthetically appealing patterns as can be seen in Fig.8. Future work will improve the proposed model to increase the classification accuracy.

The proposed approach has shown an interesting link between information theory and aesthetic. The meaning and the deeper relation of this link are to be further investigated in future work. Finding the most fundamental law or the optimization process that underlies aesthetically appealing patterns would be of great interest to the research in this area. It is interesting to see whether the proposed approach will have any link to the aesthetic judgment mechanism in the brain, and how is that related to information theory. Pursuing these research directions holds a great promise to a deeper understanding of many important phenomena.

## 5. Conclusion

A novel approach to classify aesthetically appealing images was presented in this paper. The proposed approach showed that aesthetically appealing images deliver a higher amount of information over multiple levels in comparison with less aesthetically appealing images when the same amount of energy is used. The results have shown that the proposed approach was able to classify aesthetically appealing patterns. Future work will try to apply this approach to other types of images.

## References

1. A. Chatterjee, "Neuroaesthetics: a coming of age story," Journal of Cognitive Neuroscience, vol. 23, no. 1, pp. 53–62, 2011.
2. H. Leder, B. Belke, A. Oeberst, and D. Augustin, "A model of aesthetic appreciation and aesthetic judgments," British Journal of Psychology, vol. 95, no. 4, pp. 489-508, 2004.
3. K. Hammermeister, "The German aesthetic tradition," Cambridge University Press, 2002.
4. T. Gracyk, "Hume's aesthetics," Stanford encyclopedia of Philosophy, winter 2011.
5. D. Burnham, "Kant's aesthetics" Internet encyclopedia of philosophy, 2001.
6. J. Shelley, "The concept of the aesthetic," Stanford encyclopedia of Philosophy, spring 2012.
7. E. A.Vessel, and N. Rubin, "Beauty and the beholder: highly individual taste for abstract but not real-world images," Journal of Vision, vol. 10, no. 2, 2010.
8. J. McCormack, "Facing the future: Evolutionary possibilities for human- machine creativity," Springer, pp. 417–451, 2008.
9. W.H. Latham, S. Todd, "Computer sculpture," IBM Systems Journal, vol. 28, no. 4, pp. 682–688, 1989.
10. R. Datta, D. Joshi, J. Li, and JZ. Wang, "Studying aesthetics in photographic images using a computational approach," ECCV, 2006.
11. Y. Ke, X. Tang, and F. Jing, "The design of high-level features for photo quality assessment," CVPR, pp. 288-301, 2006.
12. TO. Aydın, A. Smolic, and M. Gross, "Automated aesthetic analysis of photographic images," IEEE transactions on visualization and computer graphics, vol. 21, no. 1, pp. 31-42, 2015.
13. S. Bhattacahrya, R. Sukthankar, and M. Shah, "A framework for photo-quality assessment and enhancement based on visual aesthetics," in Proc. of the international conference on Multimedia, pp. 271–280, 2010.
14. Y.-J. Liu, X. Luo, Y.-M. Xuan, W.-F. Chen, and X.-L. Fu, "Image retargeting quality assessment," Computer Graphics Forum (Proc. of Eurographics), vol. 30, no. 2, pp. 583–592, 2011.
15. L. Liu, Y. Jin, and Q. Wu, "Realtime aesthetic image retargeting," pp. 1–8, 2010.
16. L. Liu, R. Chen, L. Wolf, and D. Cohen-Or, "Optimizing photo composition," Computer Graphics Forum, vol. 29, 2010.
17. P. O'Donovan, A. Agarwala, and A. Hertzmann, "Color compatibility from large datasets," ACM Transactions on Graphics (Proc. of SIGGRAPH), vol. 30, no. 4, 2011.
18. D. Cohen-Or, O. Sorkine, R. Gal, T. Leyvand, and Y.-Q. Xu, "Color harmonization," SIGGRAPH, pp. 624–630, 2006.
19. M. Nishiyama, T. Okabe, I. Sato, and Y. Sato, "Aesthetic quality classification of photographs based on color harmony," in Proc. of CVPR, pp. 33–40, 2011.
20. S. Dhar, V. Ordonez, and T. L. Berg, "High level describable attributes for predicting aesthetics and interestingness," in IEEE Conference on Computer Vision and Pattern Recognition (CVPR), pp. 1657–1664, 2011.
21. X. Lu, Z. Lin, H. Jin, J. Yang, and J. Z. Wang, "Rapid: Rating pictorial aesthetics using deep learning," in Proc. ACM Int. Conf. Multimedia, pp. 457–466, 2014.
22. Y. Kao, C. Wang, and K. Huang, "Visual aesthetic quality assessment with a regression model," in Proc. IEEE Int. Conf. Image Process., pp. 1583 – 1587, 2015.
23. X. Lu, Z. Lin, X. Shen, R. Mech, and J. Z. Wang, "Deep multi-patch aggregation network for image style, aesthetics, and quality estimation," in Proc. IEEE Int. Conf. Comput. Vis., pp. 990–998, 2015.
24. L. Mai, H. Jin, and F. Liu, "Composition-preserving deep photo aesthet- ics assessment," in Proc. IEEE Conf. Comput. Vis. Pattern Recognit., pp. 497–506, 2016.
25. G. Birkhoff, "Aesthetic Measure," Harvard University Press, 1933.
26. H.J. Eysenck, "An experimental study of aesthetic preference for polygonal figures," The Journal of General Psychology, vol. 79, no. 1, pp. 3–17, 1968.
27. H.J. Eysenck, "The empirical determination of an aesthetic formula," Psychological Review, vol. 48, no. 1, 1941.
28. H.J. Eysenck, "The experimental study of the 'good gestalt' a new approach," Psychological Review, vol. 49, no. 4, 1942.
29. MA. Javid, T. Blackwell, R. Zimmer, and MM. Al-Rifaie, "Correlation between Human Aesthetic Judgement and Spatial Complexity Measure," International Conference on Evolutionary and Biologically Inspired Music and Art, pp. 79-91, Mar, 2016.
30. H.W. Franke, "A cybernetic approach to aesthetics," Leonardo, vol. 10, no. 3, pp. 203–206, 1977.
31. N. Murray, L. Marchesotti, and F. Perronnin, "AVA: A large-scale database for aesthetic visual analysis," IEEE conference on Computer Vision and Pattern Recognition (CVPR), pp. 2408-2415, Jun 2012.
32. Y. Ke, X. Tang, and F. Jing, "The design of high-level features for photo quality assessment," In CVPR, 2006.
33. R. Datta, D. Joshi, J. Li, and J. Z. Wang, "Studying aesthetics in photographic images using a computational approach," In ECCV, pp. 7–13, 2006.
34. T. D. Muller, P. Clough, and B. Caput, "Experimental evaluation in visual information retrieval," the information retrieval series, Springer, 2010.
35. W. Luo, X. Wang, and X. Tang, "Content-based photo quality assessment," In ICCV, 2011.
36. A. M. Khalili, "On the mathematics of beauty: beautiful images," arXiv preprint arXiv:1705.08244v5, 2017.
37. L. Boltzmann, "Über die Beziehung zwischen dem zweiten Hauptsatz der mechanischen Wärmetheorie und der Wahrscheinlichkeitsrechnung respektive den Sätzen über das Wärmegleichgewicht." Sitzungsberichte der Kaiserlichen Akademie der Wissenschaften in Wien, Mathematisch-Naturwissenschaftliche Classe. Abt. II, 76, 1877, pp. 373–435. Reprinted in Wissenschaftliche Abhandlungen, vol. II, p. 164–223, Leipzig: Barth, 1909.
38. J.C. Maxwell, "Illustrations of the dynamical theory of gases. Part I. On the motions and collisions of perfectly elastic spheres," The London, Edinburgh and Dublin Philosophical Magazine and Journal of Science, 4th Series, vol.19, pp.19-32, 1860.
39. J.C. Maxwell, "Illustrations of the dynamical theory of gases. Part II. On the process of diffusion of two or more kinds of moving particles among one another," The London, Edinburgh and Dublin Philosophical Magazine and Journal of Science, 4th Series, vol.20, pp.21-37, 1860.
40. J. Rigau, M. Feixax, M. Sbert, "Informational aesthetics measures," IEEE Computer Graphics and Applications, vol. 28, no. 2, 2008.
41. M. R. V. Sahyun, "Aesthetics and entropy III. Aesthetic measures", 2018.
42. E. Heijer, and A. E. Eiben, "Using aesthetic measures to evolve art," Evolutionary Computation (CEC), IEEE Congress on, pp. 1-8, 2010.
43. T. Jacobsen, "Beauty and the brain: culture, history and individual differences in aesthetic appreciation," Journal of anatomy, vol. 216, no. 2, pp. 184–191, 2010.
44. T. Jacobsen, L. Hofel, "Aesthetic judgments of novel graphic patterns: analyses of individual judgments," Perceptual and motor skills, vol. 95, no. 3, pp. 755–766, 2002.
45. A. Gartus, H. Leder, "The small step towards asymmetry: Aesthetic judgment of broken symmetries," i-Perception, vol. 4, pp. 361–364, 2013.
46. L. Hofel, T. Jacobsen, "Electrophysiological indices of processing symmetry and aesthetics: A result of judgment categorization or judgment report?," Journal of Psychophysiology, vol. 21, no. 1, pp. 9-21, 2007.
47. P. P. L. Tinio, H. Leder, "Just how stable are aesthetic features? Symmetry, complexity and the jaws of massive familiarization," Acta Psychologica, vol. 130, pp. 241–250, 2009.
48. P. P. L. Tinio, A. Gartus, H. Leder, "Birds of a feather...Generalization of facial structures following massive

familiarization," Acta Psychologica, vol. 144, no. 3, pp. 463–471, 2013.